# Learned versus Kinematic Trajectory Prediction under Reduced Observation Frequency

Nukhbala Gadzhiev[1*], Alexander Morochko[1]

[1] Department of Mathematical Cybernetics and Information Technologies, Faculty of Information Technologies, Moscow Technical University of Communications and Informatics, 8A Aviamotornaya St., 111024 Moscow, Russia

[*] Corresponding author, e-mail: nleaxy@gmail.com

**Abstract**

Trajectory prediction models for automated driving are usually trained on data with a fixed sampling frequency, while positions received via vehicle-to-everything (V2X) communication arrive at 1–10 Hz. We study whether a learned predictor that uses only the observation history keeps its advantage over constant velocity (CV) and constant acceleration (CA) models when the history frequency drops. On the Argoverse 2 motion forecasting dataset, the 5 s history of the focal agent was downsampled from 10 Hz to 5, 2, and 1 Hz, and the 6 s future was predicted without maps or neighboring agents. An LSTM-based recurrent neural network (RNN) trained for each frequency was compared with the same network trained at 10 Hz and applied to sparse histories after linear interpolation. Hypotheses and thresholds were fixed before training, the main RNNs were trained with ten seeds each, and results on 24,984 test scenarios have bootstrap confidence intervals over scenarios and seeds. The average displacement error of the frequency-specific RNN grew by 3.1% at 5 Hz, 10.1% at 2 Hz, and 20.3% at 1 Hz. Contrary to our hypothesis, its advantage over the better kinematic model did not vanish on straight motion and was 24–30% on both straight motion and turns at every frequency, while for pedestrians CV was more accurate. The 10 Hz model with interpolation was at most 1.3% less accurate than the frequency-specific model down to 2 Hz but 11.5% less accurate at 1 Hz. Each hypothesis was only partly supported.

## 1 Introduction

Autonomous vehicles need to know where surrounding road users will be in the next few seconds: the trajectory prediction module takes the recent history of an object's positions and passes the predicted future positions to the motion planner (Abdel Madjid et al., 2026). When this history comes from other road users via vehicle-to-everything (V2X) communication, its rate varies: Cooperative Awareness Messages (CAMs) are sent between 1 and 10 times per second depending on vehicle dynamics, and less often when the radio channel is congested (ETSI, 2019). Prediction models, in contrast, are usually trained at a fixed sampling rate, such as 10 Hz in Argoverse 2 (B. Wilson et al., 2021).

A lower message rate is known to reduce how accurately a vehicle knows the current positions of its neighbors (Schmidt et al., 2010), and Liu et al. (2026) studied sampling frequency for end-to-end driving models, where it controls how many training samples are taken from each recording. Neither line of work measures how prediction several seconds ahead depends on the number of observations in the input history.

The most direct evidence comes from STEP, a platform for benchmarking trajectory prediction models (Schumann et al., 2025). On Argoverse 2, its authors varied the time step of the history between 0.1 and 0.5 s (10 to 2 Hz) for four state-of-the-art models, and the densest input often gave the poorest results. It was one of several checks, with one result per setting, although the same study showed that training noise can make small differences between models insignificant. The models also used maps and neighboring agents, so the role of the observation history itself remains unclear.

Two further questions remain open. First, kinematic models, such as constant velocity (CV) and constant acceleration (CA), extend the current motion of an object, and CV is a strong baseline that outperformed several neural network methods for pedestrians (Schöller et al., 2020). Whether a learned model keeps its advantage over them at low frequencies, where sparse points also make kinematic estimates of velocity and acceleration less reliable, and whether this differs between road users and maneuvers, is not known.

Second, the rate of incoming data changes while the vehicle is driving, so a model cannot be retrained for every rate. Models lose accuracy when the history length at test time differs from training (Xu and Fu, 2024), also for shorter histories on Argoverse 2 (Fan et al., 2026), and pre-trained human trajectory models often need fine-tuning at a new observation frequency (Gao et al., 2025).

To our knowledge, it has not been measured how much accuracy a vehicle trajectory model that uses only the observation history loses at lower frequencies without retraining, and whether restoring the 10 Hz grid reduces this loss.

We follow up on the frequency experiment of Schumann et al. (2025) with the observation history as the only input. On Argoverse 2, the 5 s history of the focal agent is downsampled from 10 Hz to 5, 2, and 1 Hz, without maps or neighboring agents. We compare the CV and CA models, a recurrent neural network (RNN) trained separately for each frequency, and the network trained at 10 Hz and applied to sparse data after linear interpolation, with several seeds per network and confidence intervals for all differences.

Our aim is to determine whether, and under which conditions, a learned trajectory predictor keeps its advantage over kinematic models when the observation frequency decreases. ADE is the primary metric, all thresholds were fixed before the experiment, and a relative difference below 5% is treated as practically negligible.

- RQ1. How does the accuracy of a learned model that uses only the observation history change when the frequency decreases from 10 to 1 Hz? H1: reducing the frequency from 10 to 5 Hz does not increase the ADE of the frequency-specific RNN by more than 5%, in line with Schumann et al. (2025), while at 1 Hz the ADE is more than 20% higher than at 10 Hz. The 2 Hz condition is exploratory.
- RQ2. At what frequency does the learned model lose its advantage over kinematic models, and does this differ between straight-line motion and turns? H2: at 2 Hz and 1 Hz, the RNN's advantage over the better kinematic model falls below 5% on straight-line motion, but stays above 5% on turns.
- RQ3. Can a model trained at 10 Hz be used on sparser data without retraining? H3: at 2 Hz and 1 Hz, the RNN trained at 10 Hz has an ADE more than 5% higher than the RNN trained at the matching frequency.

## 2 Materials and methods

### 2.1 Data

We use the Argoverse 2 Motion Forecasting dataset (B. Wilson et al., 2021) of about 250,000 urban driving scenarios from six US cities. Each scenario lasts 11 s at 10 Hz: 5 s of observation history (50 points) and 6 s of future (60 points). We predict only the trajectory of the focal agent and do not use maps.

The test set is the official test split (24,984 scenarios), whose future trajectories, first held out for the evaluation server (B. Wilson et al., 2021), were later released by the dataset authors (Argoverse, n.d.); we downloaded them on October 26, 2026. The official validation split is not used. From the official training split, 60,000 scenarios were drawn with random seed 42 and, after filtering, shuffled with the same seed into a validation subset of 5,000, used for early stopping, hyperparameter selection, and every other choice, and a training subset of 50,000. The test set is not used for any decision, and all splits were fixed before any data were processed.

We keep only scenarios in which the focal agent is observed at all 110 time steps, so all frequency conditions use the same scenarios; none was excluded. The test set contains 22,932 vehicles (including buses), 1,589 pedestrians, and 463 two-wheelers (cyclists and motorcyclists), and 18,518 straight, 3,213 turning, 2,415 intermediate, 462 stationary, and 376 undefined scenarios (Section 2.6), so every group has at least 300 test scenarios.

### 2.2 Frequency conditions

The history is downsampled by an integer factor $k$ without low-pass filtering: counting back from the last observed point ($t$ = 0), every $k$-th point is kept, so the $n$ retained points lie at $t = -m \cdot k \cdot 0.1$ s for $m = 0, 1, \ldots, n - 1$. In the main mode, all conditions use the 5 s window: 50, 25, 10, and 5 points at 10, 5, 2, and 1 Hz ($k$ = 1, 2, 5, and 10), with spans from the first to the last retained point of 4.9, 4.8, 4.5, and 4.0 s. The 60 future points at $t$ = 0.1, …, 6.0 s are never changed.

In the additional mode, the last five points are used, so the number of points is fixed while the span grows to 0.4, 0.8, 2.0, and 4.0 s; at 1 Hz both modes coincide. A fixed number of 50 points would need a span of 49 s at 1 Hz.

### 2.3 Models

The CV model estimates velocity from the last two observations by a finite difference, and the CA model estimates velocity and acceleration from the last three; both keep them constant over the horizon, so their predictions are the same in both modes.

The frequency-specific RNN is a one-layer LSTM encoder whose last hidden state feeds a linear decoder that outputs all 60 future positions at once; this small model trains on a CPU and accepts input sequences of any length. Each retained point is described by four input features: $x$, $y$, $\Delta x$, and $\Delta y$. Here $x$ and $y$ are its coordinates in meters in an agent-centered frame, with the origin at the position at $t$ = 0, the $x$-axis along the agent's heading at $t$ = 0, and the $y$-axis pointing to the left of it. $\Delta x$ and $\Delta y$ are the spatial increments from the previous retained point, also in meters, and are zero for the first point. They are not divided by the time interval between the points. The heading is the value reported in the dataset for the last observed point, which V2X messages also include (ETSI, 2019); it is used only for the rotation, is not an input feature, and does not depend on the frequency, because the point at $t = 0$ is kept in every condition. A separate RNN is trained for each frequency and mode, except at 1 Hz, where both modes share the same models.

The RNN trained at 10 Hz, the 10 Hz model, is also applied to downsampled histories without retraining, after linear interpolation of the retained points back to its 0.1 s grid. A mini-ablation compares zero-order hold, in which each missing grid point repeats the most recent retained point, and naive input, in which the 25, 10, or 5 points are fed as they are. When the grid is restored, the grid points before the first retained point take its value, and $\Delta x$ and $\Delta y$ are computed from the restored sequence.

The time interval $\Delta t$ between retained points is not an input of any model. The frequency-specific models do not need it, because $\Delta t$ is constant in their training and test data. For the 10 Hz model, the naive input therefore changes two things at once: the grid is not restored, and consecutive points are $k \cdot 0.1$ s apart instead of the 0.1 s seen in training, so the model reads increments that cover $k \cdot 0.1$ s as if they covered 0.1 s. The result of this variant measures the combined effect of both changes.

### 2.4 Training protocol

All RNNs get the same training budget: Adam, batches of 256 scenarios, at most 50 epochs, and early stopping with a patience of 5 epochs on the validation subset, applied from the 11th epoch. Training minimizes the batch ADE, and the weights of the epoch with the lowest validation ADE are kept. Input features are standardized over the training histories of the same frequency and mode (the 10 Hz model keeps its own statistics), and targets stay in meters. Every scenario gives one sample, so, unlike in Liu et al. (2026), all variants see the same data. For each frequency, the grid hidden size {64, 128} × learning rate {1e-3, 3e-4} is searched on the validation subset with seed 0, which is not used again, and the chosen configuration is trained with seeds 1–5.

By a rule fixed in advance, if the standard deviation of validation ADE over seeds 1–5 exceeds 2% of the mean at any frequency of the main mode, all four frequencies are also trained with seeds 6–10. The grid selected a hidden size of 128 and a learning rate of 1e-3 at every frequency, and the standard deviation was 0.13%, 1.29%, 2.03%, and 1.48% of the mean at 10, 5, 2, and 1 Hz, so seeds 6–10 were added. The additional mode at 10, 5, and 2 Hz reuses the main-mode hyperparameters with seeds 1–3. In total, 65 networks were trained: 16 for the grid, 40 in the main mode, and 9 in the additional mode.

### 2.5 Metrics

ADE, the primary metric, is the mean Euclidean distance between predicted and true positions over the 60 future points, FDE is this distance at 6 s, and MR@2m is the share of predictions with FDE above 2 m, following the Argoverse convention (B. Wilson et al., 2021). The hypotheses use three relative quantities, computed on the same test scenarios:

$$\Delta(f) = \frac{\mathrm{ADE}_{\mathrm{RNN}}(f) - \mathrm{ADE}_{\mathrm{RNN}}(10)}{\mathrm{ADE}_{\mathrm{RNN}}(10)}, \quad (1)$$

$$A(f) = \frac{\mathrm{ADE}_{kin}(f) - \mathrm{ADE}_{\mathrm{RNN}}(f)}{\mathrm{ADE}_{kin}(f)}, \quad (2)$$

$$E(f) = \frac{\mathrm{ADE}_{10\to f}(f) - \mathrm{ADE}_{\mathrm{RNN}}(f)}{\mathrm{ADE}_{\mathrm{RNN}}(f)}. \quad (3)$$

Here $f$ is the frequency in Hz, RNN is the frequency-specific model, $kin$ is the better of CV and CA (chosen on the validation subset for each group and frequency), and $10 \rightarrow f$ is the 10 Hz model applied to data at frequency $f$. $\Delta$ is the error growth, $A$ is the advantage over kinematics, and $E$ is the extra error of using the 10 Hz model.

### 2.6 Motion types

Motion types are assigned once from the 10 Hz ground truth, so the groups are the same at every frequency; because they use the future, they serve only to analyze errors after the fact. Groups are assigned in a fixed order. An agent is first labeled stationary if its true future path is shorter than 1 m. Among the remaining agents, motion is undefined if no future point lies at least 2 m from the final position (mostly agents that stop with small position jitter). For all other agents, the heading change $\Delta\psi$ (0° to 180°) is the angle between the dataset heading at $t = 0$, also used to rotate the inputs, and the final direction, which points from the last future point at least 2 m from the final position to this position. Motion is straight if $\Delta\psi < 15°$ and a turn if $\Delta\psi > 45°$; intermediate (15–45°), stationary, and undefined motion are not used in H2.

For quality control, 100 random test trajectories per motion type were plotted, screened with the help of an AI assistant, and plotted again in meters where unclear. An earlier rule, which took the final direction from the last 0.5 m of the path, assigned stopped vehicles with backward position drift to turns and was replaced before any model was trained. With the final rule, 3 of the 500 trajectories are assigned wrongly, all of them turns whose dataset heading differs from the direction of motion (3 of 100, 95% Wilson interval 1.0–8.5%; E. B. Wilson, 1927), and the other groups had no errors (0 of 100 each, 95% interval 0–3.7%). H2 is also re-evaluated with thresholds of 10°/30° and 20°/60°.

### 2.7 Statistical analysis

All models are evaluated on the same test scenarios, so all comparisons are paired. The intervals cover the variation over scenarios and seeds: in each of 10,000 resamples of a two-way bootstrap (Owen, 2007), one set of test scenarios is drawn with replacement for all models, and seeds are drawn with replacement separately for the RNN of each frequency and mode (the 10 Hz model applied to sparse data reuses the 10 Hz draws). $\Delta$, $A$, and $E$ are computed from the ADE averaged over the drawn seeds and scenarios, with percentile intervals (Efron and Tibshirani, 1993).

The eight primary comparisons are fixed in advance: H1 at 5 and 1 Hz, H2 for straight motion and turns at 2 and 1 Hz, and H3 at 2 and 1 Hz, each for all road user types together, with intervals Bonferroni-adjusted to the $1 - 0.05/8 \approx 99.4\%$ level. A part of a hypothesis is supported if its interval lies entirely on the stated side of the threshold, contradicted if it lies entirely on the other side, and inconclusive otherwise; a hypothesis is accepted only if all of its parts are supported and is otherwise partly supported (at least one part supported) or not supported. Other intervals are 95% and descriptive, and groups with fewer than 300 test scenarios would get no decisions.

The minimum sufficient frequency, a separate applied criterion, is the lowest tested frequency $f$ at which the upper 95% bound of $\Delta$ is below 5%, both at $f$ and at every tested frequency between $f$ and 10 Hz.

The hypotheses, thresholds, data splits, and this analysis plan were committed to the code repository on October 14, 2026 (commit 59e0891), before any data were processed. A revised version, with the changes listed in PLAN.md, was committed on October 26, 2026 (commit 82ce880), also before any data were processed. A third version was committed on October 27, 2026 (commit 4984b3b), after data processing and before any model was trained. It replaces the motion type rule (Section 2.6) and fixes in advance which settings may be reduced if the training time does not fit the schedule. A fourth version was committed on October 29, 2026 (commit 022fc43), after a short timing run and before the main runs. With positions as the only input, the 10 Hz model did not learn to use the velocity of the agent (validation ADE 12.3 m, against 4.3 m for the constant velocity model). The spatial increments $\Delta x$ and $\Delta y$ between retained points were therefore added as inputs, input standardization was made specific to each frequency and mode, and early stopping was disabled for the first 10 epochs. The test set was not used for this decision. All versions remain in the repository history.

## 3 Results

Results are for the test set and the main mode unless stated otherwise, and the 10 Hz model receives linearly interpolated input. Values in square brackets are 95% confidence intervals (CIs); the primary comparisons use Bonferroni-adjusted 99.4% CIs, always marked as such.

### 3.1 Accuracy across frequencies

Table 1 gives ADE, FDE, and MR@2m for all models. The ADE of the frequency-specific RNN grew from 3.11 m at 10 Hz to 3.21, 3.43, and 3.75 m at 5, 2, and 1 Hz, an error growth $\Delta$ of 3.1% [2.5, 3.9], 10.1% [8.6, 11.8], and 20.3% [18.9, 21.7], and FDE and MR@2m changed in the same direction. Between seeds, the standard deviation of its test ADE was 0.4–2.3% of the mean. CV had a lower validation ADE than CA at every frequency and is the better kinematic model in the overall comparisons; its test ADE grew from 4.40 to 5.09 m, while that of CA fell from 10.24 to 5.98 m.

**Table 1** Test-set accuracy in the main mode (24,984 scenarios), means with 95% CIs. The 10 Hz model receives sparse histories after linear interpolation; at 10 Hz it is the same network as the frequency-specific RNN.

| Metric | Model | 10 Hz | 5 Hz | 2 Hz | 1 Hz |
|---|---|---|---|---|---|
| ADE, m | CV | 4.40 [4.35, 4.44] | 4.47 [4.42, 4.52] | 4.71 [4.66, 4.75] | 5.09 [5.04, 5.14] |
| | CA | 10.24 [9.96, 10.51] | 8.05 [7.89, 8.21] | 6.68 [6.57, 6.79] | 5.98 [5.90, 6.06] |
| | RNN (frequency-specific) | 3.11 [3.08, 3.15] | 3.21 [3.17, 3.25] | 3.43 [3.37, 3.49] | 3.75 [3.69, 3.80] |
| | 10 Hz model (interpolation) | – | 3.12 [3.09, 3.16] | 3.47 [3.43, 3.52] | 4.18 [4.12, 4.23] |
| FDE, m | CV | 11.62 [11.50, 11.74] | 11.74 [11.62, 11.87] | 12.16 [12.03, 12.28] | 12.84 [12.71, 12.97] |
| | CA | 30.44 [29.65, 31.22] | 23.90 [23.45, 24.35] | 19.48 [19.19, 19.79] | 17.08 [16.87, 17.29] |
| | RNN (frequency-specific) | 8.25 [8.15, 8.34] | 8.45 [8.35, 8.56] | 8.88 [8.74, 9.03] | 9.49 [9.37, 9.62] |
| | 10 Hz model (interpolation) | – | 8.25 [8.16, 8.35] | 8.85 [8.74, 8.96] | 10.27 [10.15, 10.38] |
| MR@2m, % | CV | 85.2 [84.7, 85.6] | 85.3 [84.8, 85.7] | 85.8 [85.4, 86.3] | 86.5 [86.0, 86.9] |
| | CA | 97.1 [96.9, 97.3] | 96.5 [96.2, 96.7] | 95.5 [95.3, 95.8] | 94.8 [94.5, 95.0] |
| | RNN (frequency-specific) | 80.6 [80.1, 81.2] | 80.9 [80.5, 81.4] | 82.5 [82.0, 83.1] | 84.0 [83.5, 84.5] |
| | 10 Hz model (interpolation) | – | 80.6 [80.1, 81.1] | 83.0 [82.5, 83.5] | 86.9 [86.4, 87.3] |

The minimum sufficient frequency was 5 Hz, overall and for each road user type: the upper 95% bound of Δ at 5 Hz was 3.9% overall and 4.0%, 4.8%, and 3.3% for vehicles, pedestrians, and two-wheelers, and above 5% at 2 Hz (11.8% overall). This refers to the tested frequencies, the RNN, and the data of this study.

### 3.2 Hypotheses

Table 2 lists the eight primary comparisons.

**Table 2** Primary comparisons; values and Bonferroni-adjusted 99.4% CIs are in %. A part is supported if its interval lies on the expected side of the threshold, contradicted if it lies on the other side, and inconclusive otherwise.

| Hypothesis | Group | Expected | Value | 99.4% CI | Decision |
|---|---|---|---|---|---|
| H1 (Δ), 5 Hz | all | < 5% | 3.1 | [2.2, 4.2] | supported |
| H1 (Δ), 1 Hz | all | > 20% | 20.3 | [18.4, 22.3] | inconclusive |
| H2 ($A$), 2 Hz | straight | < 5% | 27.1 | [25.2, 28.7] | contradicted |
| H2 ($A$), 1 Hz | straight | < 5% | 26.7 | [25.1, 28.2] | contradicted |
| H2 ($A$), 2 Hz | turn | > 5% | 24.4 | [21.7, 26.9] | supported |
| H2 ($A$), 1 Hz | turn | > 5% | 26.6 | [24.3, 28.7] | supported |
| H3 ($E$), 2 Hz | all | > 5% | 1.3 | [−1.0, 3.4] | contradicted |
| H3 ($E$), 1 Hz | all | > 5% | 11.5 | [9.5, 13.6] | supported |

*H1.* At 5 Hz, Δ was 3.1% (99.4% CI [2.2, 4.2]), below the 5% threshold, so this part of H1 is supported. At 1 Hz, Δ was 20.3% (99.4% CI [18.4, 22.3]): the error growth is clear, but the interval contains the 20% threshold, so this part is inconclusive, which is neither a confirmation of H1 nor evidence that the frequency has no effect.

*H2.* On straight motion, the advantage $A$ of the RNN over the better kinematic model was 27.1% at 2 Hz and 26.7% at 1 Hz, with both 99.4% CIs above 5%, which contradicts the straight-motion part of H2; on turns, $A$ was 24.4% and 26.6%, also above 5%, so the turn part is supported. Across all frequencies, $A$ stayed at 26.7–29.5% on straight motion and 24.4–26.9% on turns (Fig. 1). For turns at 1 Hz, CA was selected on the 648 validation turns (8.72 against 8.93 m for CV) but was worse on the test set (9.13 against 8.50 m); with CV as the reference, $A$ would be 21.1%. With thresholds of 10°/30° and 20°/60°, every part of H2 got the same decision, and the lowest 99.4% bound of $A$ on turns was 17.7%.

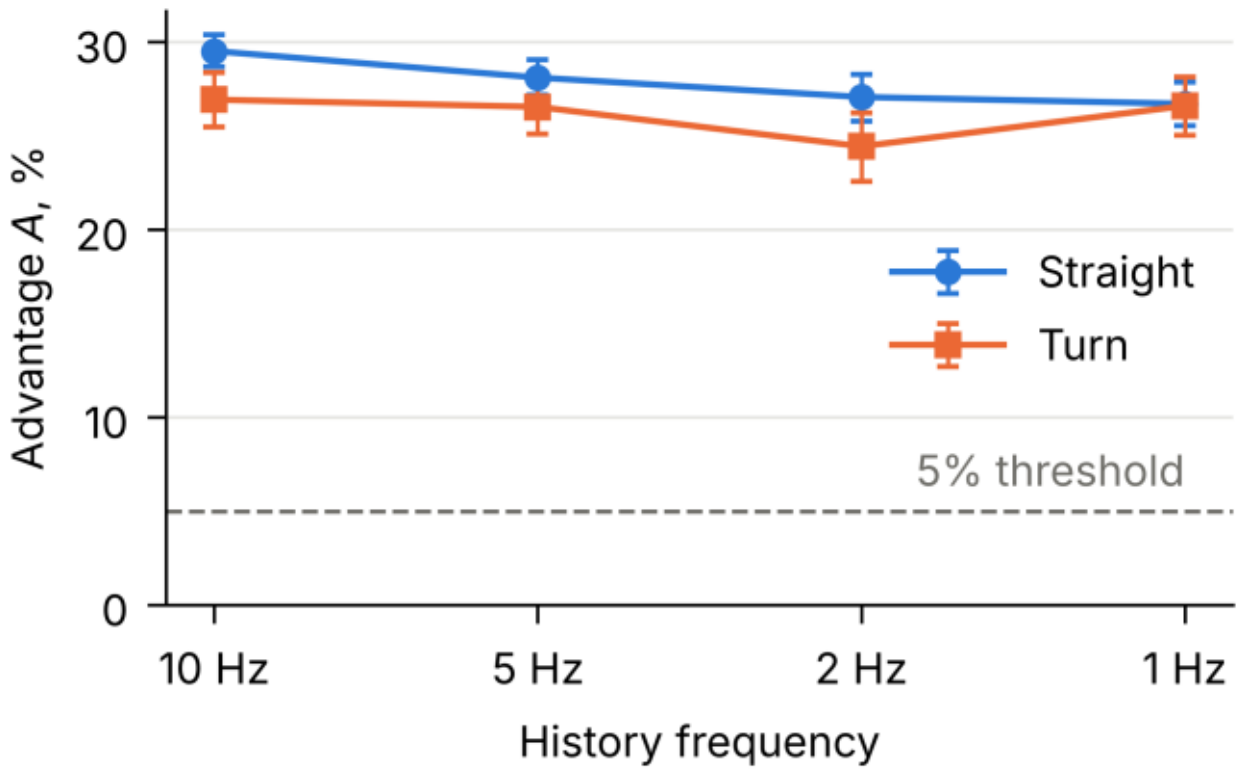

**Fig. 1** Advantage $A$ of the frequency-specific RNN over the better kinematic model on straight motion ($\Delta\psi < 15°$, 18,518 test scenarios) and turns ($\Delta\psi > 45°$, 3,213), with 95% CIs. The dashed line is the 5% threshold of H2. The kinematic model is CV, except for turns at 1 Hz, where CA was selected on the validation subset.

*H3.* At 2 Hz, the 10 Hz model had an ADE of 3.47 m, against 3.43 m for the 2 Hz RNN ($E$ = 1.3%, 99.4% CI [−1.0, 3.4]), which contradicts this part of H3. At 1 Hz, $E$ was 11.5% (99.4% CI [9.5, 13.6]; 4.18 against 3.75 m), so this part is supported. The 5 Hz condition is not part of H3; there, the 10 Hz model was more accurate than the 5 Hz RNN (3.12 against 3.21 m; $E$ = −2.7% [−3.4, −2.1]). Each hypothesis is therefore partly supported, and none is supported as a whole.

### 3.3 Restoring the 10 Hz grid

With zero-order hold, the 10 Hz model had an ADE of 7.44, 14.89, and 15.12 m at 5, 2, and 1 Hz, and with naive input 17.26, 34.02, and 38.19 m, all above the ADE of CV at the same frequency, while linear interpolation gave 3.12, 3.47, and 4.18 m. The standard deviation of ADE between seeds was 0.82–1.44 m for zero-order hold and 0.53–3.14 m for naive input, against 0.01–0.05 m for linear interpolation.

### 3.4 Road user and motion groups

CV was the better kinematic model for every road user type. For vehicles, the RNN had a lower ADE than CV at every frequency ($A$ = 30.0% [29.3, 30.6] at 10 Hz and 27.2% [26.3, 28.1] at 1 Hz), while for pedestrians it had a higher one: $A$ fell from −7.6% [−12.8, −2.5] at 10 Hz to −24.4% [−31.7, −17.8] at 1 Hz, and the ADE of CV stayed between 0.77 and 0.80 m. For two-wheelers, every interval of $A$ contains zero. The extra error $E$ at 2 and 1 Hz was 1.5% [−0.2, 3.1] and 12.0% [10.5, 13.5] for vehicles and −6.8% [−11.2, −2.4] and −5.9% [−11.5, −0.4] for pedestrians.

The advantage $A$ was 31.8–32.8% on intermediate motion and rose from 35.2% at 10 Hz to 47.7% at 1 Hz on undefined motion, while for stationary agents its intervals contain zero except at 1 Hz (23.9% [14.0, 32.1]). At 1 Hz, $E$ was 8.1% [6.4, 9.8] on straight motion and 10.4% [8.3, 12.4] on turns, and it was largest for agents that stop or stand still: 122% [97, 149] for stationary agents and 105% [84, 129] for undefined motion.

### 3.5 Additional mode

With the last five points as input, the ADE of the RNN was 3.62, 3.61, 3.54, and 3.75 m at 10, 5, 2, and 1 Hz, so Δ was −0.1% [−0.9, 0.8], −2.2% [−3.9, −0.3], and 3.6% [2.3, 5.0] at 5, 2, and 1 Hz, and the advantage over CV grew from 17.8% [17.1, 18.5] at 10 Hz to 26.4% [25.5, 27.3] at 1 Hz. At 10 Hz, five points gave a higher ADE than the 50 points of the main mode (3.62 against 3.11 m).

## 4 Discussion

### 4.1 Straight motion and turns

The main finding for RQ2 is the opposite of what H2 expected: on straight motion, the advantage of the RNN stayed at 26.7–29.5% at all frequencies, about as large as on turns, so replacing it with CV there would give up about 27%. The motion groups come from the future ground truth (Section 2.6), so this comparison shows where the models differ and gives no rule for choosing a model before a maneuver. One possible explanation is that our definition of straight motion limits only the change of heading: a vehicle that brakes before an intersection or starts from a stop counts as straight, CV keeps its last speed for 6 s, and the RNN can learn typical speed profiles from 50,000 training scenarios. We did not measure speed changes, so this explanation is untested.

By the Nyquist–Shannon theorem, a signal without frequencies above $f/2$ is completely determined by its samples taken at frequency $f$ (Shannon, 1949), so at 1 Hz a change of direction within less than about 2 s cannot be fully recovered, and turns might be expected to suffer most. Yet the error growth of the RNN at 1 Hz was similar on turns (21.5% [19.3, 23.8]) and on straight motion (20.3% [18.6, 22.0]), probably because our turns are defined by the heading change after $t = 0$, which is never sampled, and the history before many turns is close to straight.

### 4.2 Relation to previous work

The steady loss with lower frequency differs from STEP (Schumann et al., 2025), where the 10 Hz history gave the best result for only one of four map-based models; STEP also used six predictions per agent and one result per setting, while we used a single prediction and ten seeds. The input form may matter too: in our timing run, an RNN that received only positions did not learn to use the velocity of the agent until $\Delta x$ and $\Delta y$ were added (Section 2.7).

As in studies of history length (Xu and Fu, 2024; Fan et al., 2026) and of observation frequency in human trajectory prediction (Gao et al., 2025), the 10 Hz model lost accuracy when its test input differed from its training input: with naive input, its ADE was 17.26–38.19 m. The model receives $\Delta x$ and $\Delta y$ but not the time interval (Section 2.3), so zero-order hold turns the increments into $k - 1$ zeros and one increment about $k$ times larger than usual, and naive input makes the agent appear to move $k$ times faster. With linear interpolation, which keeps the increments equal, the loss was negligible at 5 and 2 Hz and reached 11.5% at 1 Hz, where interpolation may also turn the small position jitter of agents that stop or stand still into a smooth drift that looks like slow motion, which would explain their large extra error.

### 4.3 Kinematic baselines

CA was the only model whose error fell as the frequency decreased (from 10.24 to 5.98 m): a second-order finite difference divides position errors by the square of the time interval, so at 10 Hz small errors in the recorded positions, such as the drift of stopped vehicles (Section 2.6), become large acceleration errors that grow with the square of time over the 6 s horizon, while at 1 Hz they change the acceleration estimate 100 times less. These conclusions apply to finite-difference estimates: on the vehicles of our validation subset, CV with the mean velocity over the 5 s history had an ADE of 8.35 m at 10 Hz, against 4.65 m for our CV. This is the likely reason why the CV baseline reported for Argoverse 2 is weaker than ours (7.75 m for vehicle-like agents with a single prediction; B. Wilson et al., 2021): that paper does not define it, but in Argoverse 1, CV uses the mean velocity over the whole history (Chang et al., 2019). With the beta version of the dataset and motorcyclists in the vehicle-like group, the numbers are not directly comparable, and our stronger CV makes the reported advantage of the RNN smaller.

### 4.4 Practical use

The advantages reported here are relative: the FDE of the RNN at 6 s was 8.25–9.49 m, and 80.6–84.0% of its predictions ended more than 2 m from the true position (Table 1). Part of these errors is probably due to the single prediction and the absence of maps, but in its current form the model is a tool for comparing input conditions and is not ready for safe use in a planner. The tested range of 10–1 Hz covers all rates allowed for CAMs, whose interval lies between 0.1 and 1 s and depends on changes of heading, position, and speed and on congestion control (ETSI, 2019). A model trained at 10 Hz with linear interpolation lost at most 1.3% against frequency-specific models down to 2 Hz, so a receiver would not need a separate model for each message rate in this range. At 1 Hz, it lost 11.5%, most for agents that stop or stand still, which send CAMs least often, although our stationary group is defined by the future path and only approximates these agents. Keeping the last message, as in zero-order hold, raised ADE to 7.4–15.1 m. All of this assumes regular intervals between messages.

### 4.5 Limitations and future work

We used one dataset of challenging urban scenarios from US cities (B. Wilson et al., 2021) and one small LSTM without maps or neighboring agents, so the results do not necessarily hold for average driving, for other neural predictors, such as transformer-based models, or for map-based models. Sparse data were simulated by regular downsampling, while real V2X messages arrive at irregular intervals, with losses and latency, and at 1 Hz linear interpolation inserts nine artificial points between every two real ones, so the H3 result at this frequency refers to this procedure. The thresholds (5%, 20%, 15°/45°) were fixed before the experiment but are not an accepted standard, so all intervals are reported in full, and ten seeds give only a rough estimate of training variation. One network is trained for all road user types without the type as an input, and its loss in meters is dominated by vehicles (92% of the test set), which may explain why CV was more accurate for pedestrians, in line with Schöller et al. (2020). In 37 of the 65 runs, the lowest validation ADE came in one of the last five epochs, more often at 5 and 2 Hz (8 of 10 main-mode runs each) than at 10 Hz (2 of 10), so a larger budget would probably help the low-frequency RNNs more, which would reduce $\Delta$ and increase $E$ and $A$; the H3 decision at 2 Hz is the most sensitive, with an upper 99.4% bound of $E$ of 3.4%. The plan was revised three times before the main runs (Section 2.7). Future work could use real V2X message logs and give the time interval $\Delta t$ to the model as an input, as frame-rate conditioning does for human trajectories (Gao et al., 2025).

## 5 Conclusion

On 24,984 test scenarios of Argoverse 2, a learned predictor that uses only the observation history lost 3.1% of its accuracy at 5 Hz and 20.3% at 1 Hz, so the 5 Hz part of H1 was supported and the 1 Hz part was inconclusive. Contrary to H2, it kept an advantage of about 27% over the better kinematic model on straight motion at 2 and 1 Hz, similar to that on turns; this advantage came from vehicles, while for pedestrians CV was more accurate. A model trained at 10 Hz with linear interpolation lost 1.3% at 2 Hz and 11.5% at 1 Hz, so retraining paid off only at 1 Hz. None of the three hypotheses was supported as a whole, and the minimum sufficient frequency was 5 Hz for the tested frequencies, model, and data. All advantages reported here are relative, and the absolute errors of the RNN remain too large for direct use in a motion planner.

## Data availability

The code, configuration files, scenario lists, per-scenario errors of all models, and the scripts that produce the tables and figures are available at https://github.com/nleaxy/learned-vs-kinematic-trajectory-prediction under the MIT license. The Argoverse 2 Motion Forecasting dataset and its test annotations are available from the dataset authors under the CC BY-NC-SA 4.0 license (B. Wilson et al., 2021; Argoverse, n.d.). Trained model weights are available from the corresponding author on request.

## Use of AI tools

Claude (Anthropic) was used to help write the code and parts of the text. The authors checked all code, results and text and take full responsibility for the content.